\documentclass[print,12pt]{elsarticle}

\usepackage{amsmath}
\usepackage{stackrel,amssymb}
\usepackage[title,titletoc]{appendix}

\graphicspath{{.}{./graphs/}{./images/}{./Figures/}{./figures/}}

\title{Stochastic modeling of non-linear adsorption with Gaussian kernel density estimators}

\author[add1,add2]{Maryam Rahbaralam \corref{cor1}}
\ead{maryam.rahbaralam@upc.edu}
\cortext[cor1]{Corresponding author (maryam.rahbaralam27@gmail.com)}

\author[add3]{Amir Abdollahi}
\author[add1,add2]{Daniel Fern\`{a}ndez-Garcia}
\author[add1,add2]{Xavier Sanchez-Vila}

\address[add1]{Department of Civil and Environmental Engineering (DECA), Universitat Politècnica de Catalunya, UPC-BarcelonaTech, Barcelona, Spain}
\address [add2] {Hydrogeology Group (GHS), UPC-CSIC, Barcelona, Spain}
\address[add3]{Laboratori de Calcul Numeric (LaCaN), Department of Civil and Environmental Engineering (DECA), Universitat Politècnica de Catalunya, UPC-BarcelonaTech, Barcelona, Spain}

\begin{document}

\begin{frontmatter}

\begin{abstract}
Adsorption is a relevant process in many fields, such as product manufacturing or pollution remediation in porous materials. Adsorption takes place at the molecular scale, amenable to be modeled by Lagrangian numerical methods.
We have proposed a chemical diffusion-reaction model for the simulation of adsorption, based on the combination of a random walk particle tracking method involving the use of Gaussian Kernel Density Estimators. 
The main feature of the proposed model is that it can effectively reproduce the nonlinear behavior characteristic of the Langmuir and Freundlich isotherms.
In the former, it is enough to add a finite number of sorption sites of homogeneous sorption properties, and to set the process as the combination of the forward and the backward reactions, 
each one of them with a prespecified reaction rate. To model the Freundlich isotherm instead, typical of low to intermediate range of solute concentrations, 
there is a need to assign a different equilibrium constant to each specific sorption site, provided they are all drawn from a truncated power-law distribution.
Both nonlinear models can be combined in a single framework to obtain a typical observed behavior for a wide range of concentration values.
\end{abstract}

\end{frontmatter}

\section{Introduction}

Adsorption is defined as the binding of atoms or molecules from a gas or liquid to a surface. It is a phenomenon well-described in many physical, biological, and chemical systems and processes, and that has been widely employed in industrial applications such as pharmaceutical industry, chillers and air conditioning systems, water purification, coatings, and resins, to name a few. To design and optimize an adsorption-based process, it is necessary to characterize accurately the adsorption equilibria and their dependence on the experimental conditions. 

Equilibrium relations are described by adsorption isotherms, relating the equilibrium concentration of a solute on the surface of an adsorbent to the concentration of the solute in the liquid/gas being in contact. In 1916, Langmuir introduced the first scientifically based nonlinear isotherm by assuming a homogeneous surface with a specific number of sites where the solute molecules could be adsorbed \citep{Langmuir1918}. Furthermore, he assumed that the adsorption involves the attachment of only one layer of molecules to the surface, i.e. mono-layer adsorption. However, the Langmuir model deviates from the experimental observations in the presence of a rough inhomogeneous surface where multiple site-types or
layers are available for adsorption and some parameters vary from site to site. This problem was tackled by Freundlich who proposed the first mathematical fit to a nonlinear isotherm, leading to a purely empirical formula for adsorption 
on heterogeneous surfaces \citep{Freundlich1906}. The Freundlich isotherm was established by assuming that adsorption varies directly with pressure without reaching saturation, while experimentally the rate of adsorption saturates by applying very high pressures. Therefore, the use of the Freundlich isotherm is appropriate when dealing with low/medium pressures/concentrations. 

The Langmuir and Freundlich models are the two most commonly used isotherms due to their simplicity and their ability to properly fit a variety of adsorption experimental data. Several isotherms have been also proposed to combine the features of both the Langmuir and Freundlich isotherms, including the BET \citep{BET}, Sips\citep{Sips1948} and Redlich-Peterson \citep{Redlich1959} models. Apart from these classical isotherms, other thermodynamically consistent adsorption models can be derived from a fundamental integral equation relating the experimental isotherm, the adsorption energy distribution, and the local isotherm, see \citet{Quinonesa} for a review of these models. 

A major challenge to control and predict an adsorption-based process is the heterogeneity and complexity of interactions between the adsorbate and adsorbent surfaces in a dynamic solid/fluid or solid/gas system. This complexity is the main reason behind proposing various isotherm models to predict an adsorption process based on macroscopic experimental data. To tackle this complexity, computational models have been developed by incorporating the interaction of the reactants into the dynamic system at the molecular level. A number of these models employ Density Functional Theory (DFT) methods, force-field techniques and Molecular Dynamics (MD) methods to simulate adsorption at the molecular level, see \citet{Costa2015} for a recent review.  These models are ideal to study adsorption close to the surface; however, due to a high computational cost, they cannot be employed for large simulation domains, i.e., at the pore scale or above.

To span over a wide range of scales, continuum models based on a diffusion-reaction equation have been developed to study reactive solute transport driven by diffusion.  This classical problem can be addressed using both Eulerian and Lagrangian approaches, taking into account their limitations and advantages. In an Eulerian approach, the problem is defined in terms of reactants concentration which is used to describe the reaction rate by macroscopic laws.  A macroscopic mass balance equation such as the diffusion-reaction equation (DRE) is then obtained in terms of the concentration of a solute $[C]$ that is in instantaneous equilibrium at each point in space with the adsorbed concentration [S], both expressed as mass per unit volume of solute. The system is then formulated in terms of two coupled equations:

\begin{equation}
\label{DRE1}
\dfrac{\partial [C]}{\partial t}=D \nabla^{2} [C]-\dfrac{\partial [S]}{\partial t}
\end{equation}

\begin{equation}
\label{DRE2}
[S]=f([C]).
\end{equation}
where $D$ is the diffusion coefficient. Alternative expressions for Eq. \eqref{DRE1} involve defined $[S]$ as mass of sorbed species per unit mass of solid. These two equations can be combined to write down a single one

 \begin{equation}
\label{DRE3}
(1+f'([C])) \dfrac{\partial [C]}{\partial t}=D \nabla^{2} [C].
\end{equation} 

Except in the particular case of Eq. \eqref{DRE2} being a linear relationship, the system results in a nonlinear partial differential equation, thus limiting the usefulness of Eulerian approaches to tackle the mathematical problem. An approach to simulate nonlinear sorption with particle tracking based on Eq.~\eqref{DRE3} has been presented by \citet{Tompson1993}. This author estimated $f'([C]))$ by reconstructing concentrations from particles at each time step
 without controlling the statistical errors involved. In this context, the method presented here does not require the computation of solute concentrations during the course of the simulation, therefore the statistical errors cannot propagate overtime. 
 
In the presence of chemical heterogeneity, small enough volumes are required to capture the reactions while a macroscopic quantity such as concentration (mass per unit volume) loses its integrity in this limit. An alternative approach is to address this problem at the molecular level within a Lagrangian framework, where the movement of each individual molecule is tracked \citep{Gillespie1976, Gillespie1977, Gillespie2000}. However, since an extremely large number of molecules can exist even at very small concentrations, this approach is computationally unfeasible for simulating reactive transport in porous media. Moreover, at the molecular level, the continuum assumption is no longer valid and chemical kinetics should be derived by molecular collision.

Particle Tracking Methods (PTM) offer a practical and efficient alternative to overcome these problems by combining the key features of both approaches. In this case, the solute plume is divided into manageable number of particles, typically ranging between $10^{6} - 10^{9}$. This way, each particle does not represent a molecule itself but a certain fraction of mass containing numerous molecules. This advantage comes with the need to translate reaction-rate equations into particle relationships. In non-linear reactive transport problems, the latter includes the chemical 
interaction between particles, which is normally defined based on the particle area of influence. A concept that has different interpretations among researchers.

Kernel Density Estimators (KDE) are typically used in statistics as a non-parametric approach to estimate probability distribution function from a finite data sample. Its application in Lagrangian reactive transport problems has several advantages. Since it is non-parametric, KDE allows the identification of complex solute plume distributions (multimodal or non-Fickian behaviors). It provides not only adequate estimates of concentrations but also of their functionals (e.g. mixing, Human-health risk). And more importantly, it provides an adequate mathematical framework to select an optimal choice of the particle area of influence \citep{Fernandez-Garcia2011}. This parameter will essentially dictate if particle of appropriate kinds are mutually in contact for the occurrence of chemical reactions. In this context, \citet{Rahbar2015} has demonstrated that this approach avoids the segregation of particles and the resulting incomplete mixing which is common in classical diffusion-based models.  A variety of kernel functions can be used, among them, Gaussian kernels are usually preferred for mathematical advantages \citep{Pedretti2013}.

Motivated by this potential, here we extend the Gaussian KDE model to simulate nonlinear adsorption. We show that the model proposed is able to reproduce the results of the Langmuir and Freundlich isotherms and to combine the features of these two classical models. This approach opens up a new way to predict and control an adsorption-based process using a particle-based method with a finite number of particles.  The paper is structured as follows. First, Section \ref{Numeric} sets out the background, the problem and the numerical approach. Section \ref{KDEmodelAD} introduces the proposed adsorption model using PTM and Gaussian KDEs. Simulation results and discussion are presented in Section \ref{Results}. Finally, Section \ref{Concl} provides a summary of the main contributions of this paper.

\section{Background and statement of the problem}\label{Numeric}

\subsection{The Langmuir isotherm}
\label{Langmuir}

The Langmuir isotherm explains adsorption by assuming that the adsorbent is an ideal solid surface composed of series of identical sites 
capable of binding the adsorbate. This binding is treated as a chemical reaction between the adsorbate molecule $A$ and an empty site, $B$.
This reaction yields an adsorbed complex $C$. This process can be reversed through desorption whereby the adsorbed molecule is released from the surface
and the complex $C$ is transformed to $A$ and $B$. This dynamic equilibrium existing between the adsorbate and the adsorbent can be expressed as

 \begin{equation}
\label{Eq2}
A+B \rightleftarrows C.
\end{equation} 

This model assumes adsorption and desorption as being elementary processes, where the rate of forward adsorption $r_f$ and the rate of backward desorption $r_b$ are given by:

 \begin{equation}
\label{rf}
r_f=k_{f}[A][B],
\end{equation} 

 \begin{equation}
\label{rb}
r_b=k_{b}[C],
\end{equation} 
where $k_{f}$ is the forward adsorption reaction constant, $k_{b}$ is the backward desorption reaction constant, and $[X]$ denotes the concentration of species $X$ ($A$, $B$, or $C$). At equilibrium, the rate of adsorption equals the rate of desorption, i.e. $r_f=r_b$, then by rearranging the terms we obtain

 \begin{equation}
\label{Keq}
\frac{[C]}{[A][B]} = \frac{k_f}{k_b} = K_{eq},
\end{equation} 
where $K_{eq}$ is the equilibrium constant. By adding up the concentration of free sites $[B]$ and of occupied sites $[C]$, the concentration of all sites $[B_0]$, assumed constant in time, is obtained as $[B_0] = [B] + [C]$. Combining this relation and Eq.~\eqref{Keq} yields the Langmuir adsorption isotherm:

 \begin{equation}
\label{Lang}
[C] = [B_0]\frac{K_{eq}[A]}{1+K_{eq}[A]}.
\end{equation} 

In the presence of a high concentration of the adsorbate $[A]$, Eq.~\eqref{Lang} leads to the saturation of surface sites, $[C] \longrightarrow [B_0]$. In other words, the surface reaches a saturation point where the maximum adsorption capacity of the surface will be achieved. 

\subsection{The Freundlich isotherm}
\label{FI}
The Freundlich isotherm is an empirical model, which is commonly used to describe the adsorption performance of heterogeneous surfaces. 
This isotherm is mathematically expressed as

 \begin{equation}
\label{Frd}
[C] = K[A]^{m},
\end{equation} 
where $K$ and $m$ are called the Freundlich constants. The constant $K$ is an adsorption coefficient, while $m$ is a measure of the deviation from the linearity of the adsorption. Unlike in the Langmuir model, in this one there is no adsorption maximum or saturation. Equation \eqref{Frd} can be linearized as

 \begin{equation}
\label{LogFrd}
\log [C] = \log K + m\log[A],
\end{equation} 
where $m$ represents the slope of the line in a $log[C]-log[A]$ plot. 

While the Freundlich model has been found to fit most existing adsorption experimental data \citep{Frankenburg1944,Davis1946,Thomas1957,Urano1981}, its theoretical basis is under scrutiny. In principle, the Freundlich isotherm can be derived from the fundamental integral equation for the overall adsorption isotherm stated as

 \begin{equation}
\label{integ}
[C] = \int_\Omega N(Q)\Theta(Q) dQ,
\end{equation} 
where $N$ is the number of sites having adsorption energy $Q$ and $\Theta$ is the local coverage of individual sites.
The integration region $\Omega$ is over all possible adsorption energies. It has been shown that Eq.~\eqref{integ} leads to
the Freundlich isotherm when an exponential distribution of adsorption energies is assumed \citep{Sips1948,Sheindorf1981}:

 \begin{equation}
\label{N}
N(Q) = m~\text{exp}(-mQ/RT),
\end{equation} 
where $m$ and $R$ are constants, and $T$ is the temperature (assumed constant as implied by the word isotherm). Furthermore, it is assumed that the local coverage follows the Langmuir isotherm as 

 \begin{equation}
\label{LangFr}
\Theta = \frac{\hat K[A]}{1+ \hat K[A]}.
\end{equation} 
with the equilibrium constant $\hat K$ depending on the adsorption energy as

 \begin{equation}
\label{K0}
\hat K = K_{eq}~\text{exp}(Q/RT).
\end{equation} 

\noindent Plugging Eqs.~\eqref{N} - \eqref{K0} into Eq.~\eqref{integ} and assuming that $Q \in (-\infty,+\infty)$ yields

 \begin{equation}
\label{integFr}
[C] = \int_{-\infty}^{+\infty} m~\text{exp}(-mQ/RT) \frac{K_{eq}~\text{exp}(Q/RT)[A]}{1+K_{eq}~\text{exp}(Q/RT)[A]}   dQ,
\end{equation} 
This integral has a solution in the form of the Freundlich isotherm given by

 \begin{equation}
\label{integF}
[C] =  \overline{K} [A]^m,
\end{equation} 
with the Freundlich constant $\overline{K} = m\pi RT {K_{eq}}^m / \sin((1-m)\pi)$ . Therefore, Eqs.~\eqref{integ}-\eqref{integF} establish a theoretical basis for 
the Freundlich isotherm. In Section \ref{KDEmodelAD}, we demonstrate that this approach, which is based on an exponential distribution of adsorption energy, is mathematically equivalent to consider $\hat K$, in the Langmuir model, as a random variable 
that follows a truncated power-law distribution. 

\subsection{The KDE-based diffusion-reaction model}
\label{KDEmodel}

This section presents a summary of our recently proposed chemical reaction model using KDEs \citep{Rahbar2015}. This model has been developed based on a single forward bimolecular irreversible reaction $A+B\rightarrow\emptyset$, where $A$ and $B$ represent two species in the dissolved phase, reacting kinetically with unitary stoichiometric reaction coefficients.  A diffusion-reaction equation governs the transport of the reactants:

 \begin{equation}
\label{A-D1}
\dfrac{\partial C_i}{\partial t}=D \nabla^{2} C_{i} - k_{f} C_{A} C_{B},
\end{equation} 
\noindent where $i$ represents $A,B$, the concentration of species $i$ is indicated by $C_{i}=C_{i}(x,t)$, $k_{f}$ is the reaction rate constant, and $D$ is the diffusion coefficient.

To simulate this problem, the random walk particle tracking method is used whereby an equal fraction of the total mass is carried by each particle. For simplicity and without loss of generality, the 1-D form of the reactive-advective-dispersive equation is considered. Given $j$ as the particle number, the locations of the A and B particles, $X_{j}^{A}$ and $X_{j}^{B}$, are initialised using a statistical uniform distribution. Then, a Brownian random walk motion governs the diffusion of the species during a time interval between $t$ and $t$+$\Delta t$, formally written as

 \begin{equation}
\label{BRW}
  X_{j}^{i} (t+\Delta t)=X_{j}^{i} (t)+\xi_{t}~\sqrt{2D\Delta t} ,~ i=A,B
\end{equation} 
\noindent where $v$ is the flow velocity, $2D \Delta t$ is the variance, and $\xi_{t}$ is a normally distributed random number drawn at each time step. 

As stated in the introduction, Kernels provide an influential area around each particle. This area is controlled by a parameter $h$ whose optimal value is obtained based on minimising the variance error while maintaining smoothness. The optimum bandwidth is found as \citep{Park1990}

 \begin{equation} \label{hopt}
h^{opt}= G~N^{-1/5},
\end{equation} 

\noindent where the value of $G$ is time dependent and can be formally derived from the second derivative of the concentration spatial function. So, from Eq.~\eqref{hopt}, the optimal bandwidth size $h^{opt}$ increases inversely proportional to the number of particles. This feature is particularly beneficial to avoid incomplete mixing due to the segregation of particles during chemical reactions. 

In a Reactive Particle Tracking (RPT) method, the reaction of particles is simulated through probabilistic rules,

 \begin{equation}
\label{Pf DB}
P_{f}=k_{f}~m_{p}~\Delta t~v(r,\Delta t),
\end{equation} 

\noindent where $m_p$ is the particle mass and $P_{f}$ is the (forward) probability that two particles, $A$ and $B$, separated by a distance $r$ react within the time interval $\Delta t$. The co-location probability density function (pdf) is represented by $v(r,\Delta t)$ which defines the probability that two particles separated by a distance $r$ occupy the same position after a time interval $\Delta t$. By attributing a Gaussian density with a standard deviation $\sigma$ to each particle, the convolution of two particles density results in the co-location pdf, representing the reaction zone of the two particles. This probabilistic rule was demonstrated by \citet{Benson2008}. The fundamental difference in the KDE model is that the area of influence, $\sigma=h^{opt}$, around a particle is not only attributed to diffusion but also depends on the distribution of particles (shape and number of particles). This dynamic area of influence changes the probability of reaction at each time step, avoiding the formation of segregated areas of particles. Based on the principles of the law of mass action, the probability of forward reaction for the KDE-based model is obtained as  \citep{Rahbar2015}

 \begin{equation}
\label{preaction}
P_{f}=\dfrac{k_{f} ~ m_p ~\Delta t}{2h^{opt}\sqrt{\pi}} ~ \exp\Big(\dfrac{-r^{2}}{(2h^{opt})^{2}}\Big).
\end{equation} 
This equation implies that by increasing the area of influential of each particle, $h^{opt}$, the probability of reaction decreases while the unitary area under the co-location pdf is preserved. Such expansion, in turn, would increase the number of potential reactive pairs.

\section{The KDE-based adsorption model}
\label{KDEmodelAD}

The probability of forward reaction $P_{f}$ provided in Eq.~\eqref{preaction} was originally proposed to model a simple chemical model, i.e., the single forward bimolecular irreversible reaction. To extend this approach for modeling adsorption, $P_{f}$ is considered as the probability of forward adsorption reaction in Eq.~\eqref{Eq2} with the particle $C$ as the product of the reaction. In contrast to the bimolecular irreversible reaction, an adsorption reaction is reversible, so that a particle $C$ can transform back to a particle $A$ and another one $B$. The probability of backward desorption is given by

 \begin{equation}
\label{Pb}
P_{b}=k_{b}\Delta t.
\end{equation} 

To numerically implement the probability of forward and backward reactions in Eqs.~\eqref{preaction} and ~\eqref{Pb}, we use the following approach. In the forward case, we first computed $P_{f}$ from Eq.~\eqref{preaction}, and then we compared this value with a random number generated from a uniform distribution $\xi_{f}\sim U(0,1)$. Then, if $P_{f}\geq\xi_{f}$, the reaction was supposed to have occurred, and both $A$ and $B$ particles were removed from the system and substituted by a $C$ particle; otherwise, the reaction was not supposed to have taken place at that particular time step and both particles were kept. This procedure was repeated for every pair of particles (A,B) in the system. 

A reversed procedure was repeated for the backward reaction by comparing a random number $\xi_{b}$ with the probability of backward reaction $P_b$ obtained from Eq.~\eqref{Pb}. If $P_{b}\geq\xi_{b}$, the reaction was supposed to have occurred, the $C$ particle was removed and substituted by an $A$ and a $B$ particles located at the same point $C$ was originally considered; otherwise, no action was taken, representing that the reaction had not occurred. After the loop for all $C$ particles were performed, the simulation continued to the next time step. We also assumed that the locations of the adsorbent particles $B$ and the reaction product particles $C$ were fixed (the location of the sorption sites did not change with time), so that Eq.~\eqref{BRW} was only applied to the displacement of the adsorbate particles $A$. At each time step, the number of remaining particles multiplied by $m_p$ and divided by the volume resulted in the concentrations $[A]$ and $[C]$. The simulation was carried out until equilibrium was clearly achieved, as indicated by the stabilization in the ratio of concentrations $[C]/[A][B]$. 

As explained later, we set up two simulation runs to assess the capacity of the model to simulate adsorption based on both the Langmuir and the Freundlich isotherms. In order to account for the former, it is just enough to set a finite number of sites $[B_0]$ and then fix the $k_f$ and $k_b$ values to honor the relationship set in Eq.~\eqref{Keq}. In the case of the Freundlich isotherm, it is necessary to incorporate an exponential distribution of adsorption energies, as given in Eq.~\eqref{N}, and to set the individual value of the equilibrium constant $\hat K$ at each site. The generation of the equilibrium constant $\hat K$ requires several considerations. As indicated in Section \ref{FI}, the assumption of $Q_{min} \rightarrow -\infty$ in Eq.~\eqref{integFr} is necessary for the validation of the Freundlich model in Eq.~\eqref{integF}. However, several authors \citep{Hill1949,ROCHA1997} have indicated that the integration limits over all possible adsorption energies in Eq.~\eqref{integ} should have a minimum cut-off to have a physical significance. To include this in our model, we consider the site-energy distribution $N$ given by Eq.~\eqref{N} as a truncated exponential distribution with cumulative distribution function given as 

\begin{equation}
\label{Cm}
F(Q) = 1-\exp\Big(-\frac{m}{RT}(Q - Q_{min})\Big), ~~  Q\in[Q_{min},+\infty),
\end{equation}
where $Q_{min}$ is the minimum energy bound. Considering Eq.~\eqref{K0}, the cumulative distribution is obtained as a function of the equilibrium constant $\hat K$,

\begin{equation}
\label{Cm2}
F(\hat K) =  1-\exp (-m \ln\dfrac{\hat K}{K_{eq}}+m \ln\dfrac{K_{min}}{K_{eq}}), ~~ \hat K\in[K_{min}, +\infty),
\end{equation}
where $K_{min} = K_{eq} \exp({Q_{min}}/{RT})$. Notice that Eq.~\eqref{Cm2} is equivalent to a truncated power-law function, given as

\begin{equation} \label{Cm33}
F(\hat K)= 1 - \Big(\dfrac{\hat K} {K_{min}}\Big)^{-m}, ~~ \hat K \in[K_{min}, +\infty),
\end{equation}
with the corresponding probability density function,

\begin{equation}\label{Cm3}
f_{\hat K}(\hat K)= \dfrac{m}{K_{min}}  \Big(\dfrac{\hat K}{K_{min}}\Big)^{-1-m},  ~~ \hat  K\in[K_{min}, +\infty).
\end{equation}

Therefore, the generation of $\hat K$ values can be determined directly from the cumulative distribution function of $F(\hat K)$. Given $\zeta$ to be a random value of a uniform distribution between 0 and 1, the corresponding $\hat K$ value is obtained as

\begin{equation}\label{Cm4}
\hat K=K _{min}  (1 - \zeta)^{-\frac{1}{m}}.
\end{equation}

Rewriting Eq.~\eqref{integ} in terms of the probability density function of equilibrium constant $\hat K$ and the concentration of the adsorbed chemical compound $C$, we have 
\begin{equation}
\label{integFrC}
[C]_a = [B_0] \int_{K_{min}}^{+\infty} f_{\hat K}(\hat K) \Theta(\hat K)  d \hat K.
\end{equation}
Introducing Eq.~\eqref{LangFr} and Eq.~\eqref{Cm3} into Eq.~\eqref{integFrC} results in
\begin{equation}
\label{integFrC2}
[C]_a = m [B_0] (K_{min})^m \int_{K_{min}}^{+\infty} \dfrac{\hat K^{-m}[A]}{1+\hat K [A]}d \hat K.
\end{equation}

This integral has two limiting solutions depending on the concentration of $[A]$. When $[A]$ is small, the solution is obtained by applying a change of 
variable in the integration as $x = \hat K [A]$, so that

\begin{equation}
\label{integFrC3}
[C]_a = m [B_0] (K_{min})^m [A]^m \int_{[A]K_{min}}^{+\infty} \dfrac{x^{-m}}{1+x}d x.
\end{equation}
Now, we can see that the lower limit of integration approaches to zero and the solution is equivalent to 

\begin{equation}
\label{integFrC4}
[C]_a = m [B_0] (K_{min})^m [A]^m \int_{0}^{+\infty} \dfrac{x^{-m}}{1+x}d x 
= m [B_0] (K_{min})^m [A]^m \frac{\pi}{\sin((1-m)\pi)}
\end{equation}
Thus, when $[A] \rightarrow 0$, the concentration of the adsorbed chemical compound $C$ follows the Freundlich isotherm written as

\begin{equation}
\label{integFrC5}
[C]_a = K[A]^m,
\end{equation}
with

\begin{equation}
\label{integFrC6}
K = \frac{m \pi [B_0](K_{min})^m}{\sin((1-m)\pi)}.
\end{equation}
On the contrary, when $[A]$ is large, Eq.~\eqref{integFrC3} simplifies to 
\begin{equation}
\label{integFrC7}
[C]_a = m [B_0] (K_{min})^m [A]^m \int_{[A]K_{min}}^{+\infty} x^{-m-1} d x,
\end{equation}
which gives

\begin{equation}
\label{integFrC8}
[C]_a = m [B_0] (K_{min})^m [A]^m \frac{[A]^{-m}(K_{min})^{-m}}{m} = [B_0].
\end{equation}
Thus, at large concentration of $[A]$, the solution approaches the solution limit as given by the Langmuir isotherm. In short,
a truncated power law distribution of equilibrium constant gives a sorption isotherm that follows the Freundlich model for small
$[A]$ concentrations and changes to the Langmuir saturation limit for large $[A]$ concentrations. To understand the transition between
the two limiting models, one can analyze the deviation of Eq.~\eqref{integFrC2} from the Freundlich model. The relative deviation from 
the Freundlich model can be written as 

\begin{equation}\label{Cm7}
\epsilon  =\frac{[C]_f  - [C]_a }{[C]_f} = \Big[\dfrac{\pi}{\sin((1-m)\pi)}\Big]^{-1}[A]^{-m}\int_{0}^{K_{min}}\dfrac{\hat K^{-m}[A]}{1+\hat K [A]}d \hat K . 
\end{equation}

To find an approximate solution of the relative deviation we expand the integral around $\hat K$=0 in an ascending series in $K_{min}$

\begin{equation}\label{series}
\int_{0}^{K_{min}}\dfrac{\hat K^{-m}[A]}{1+\hat K [A]}d \hat K=[A] K_{min}^{-m} (\dfrac{-K_{min}}{m-1}+\dfrac{[A] K_{min}^{2}}{m-2}-\dfrac{[A]^{2} K_{min}^{3}}{m-3}+...).
\end{equation}

This is a relatively good approximation given that $K_{min}$ is small and the integral limits are very close to each other. Truncating the series expansion at the first term results in

\begin{equation}\label{u-ODE}
[A]_c=\Big[\dfrac{\epsilon \pi(1-m)}{\sin((1-m)\pi) }\Big]^{{1}/{(1-m)}} K_{min}^{-1},
\end{equation}
where $[A]_c$ is the concentration of $[A]$ at which the isotherm starts deviating from the Freundlich model.

The random walk model of sorption proposed here has three parameters $[B_0]$, $m$, and $K_{min}$. The parameter $m$ represents directly the exponent of the Freundlich isotherm. The parameter $K_{min}$
determines the maximum concentration of the chemical compound $[A]$ in the liquid phase for which sorption follows the Freundlich isotherm. The parameter $[B_0]$ is the concentration of sorption sites in the 
system which relates to the Freundlich coefficient $K$ according to Eq.~\eqref{integFrC6}.

\section{Numerical simulations}
\label{Results}

\subsection{The Langmuir Model}
\label{L2}
The first test was design to show the performance of the proposed model to reproduce adsorption based on the Langmuir model. The constants of forward adsorption and background desorption were set to $k_f = 0.5$ and $k_b =0.1$, respectively. 
A 1-D domain of size $\Omega = 200$, a particle mass of $m_p = 1$, and initial concentrations of $[B_0] = 200$ and $[C_0] = 1$ are considered. 
Simulations are performed for 2000 time steps with a time interval of $\Delta t = 10^{-2}$ and a diffusion coefficient of $D = 10^{-2}$.  
All values throughout the paper are reported in normalized units.

At each time step, all the adsorbate particles $A$ are moved following Eq.~\eqref{BRW} to account for the effect of diffusion. To effectively search for potentially reactive pairs of particles $A$ and $B$, the 1-D domain
is divided by elements of size $\Omega/2h^{opt}$, then we find at each element and its two neighbors, all pairs of particles $A$ and $B$.
The distance between each pair is first obtained, then the procedure mentioned below Eq.~\eqref{Pb} is followed to implement the forward reaction. After repeating this procedure for every pair of particles 
in all elements, the backward reaction for each particle $C$ is implemented following a similar procedure mentioned in Section~\ref{KDEmodelAD}.

\begin{figure}[!htp]
   \includegraphics[width=30pc]{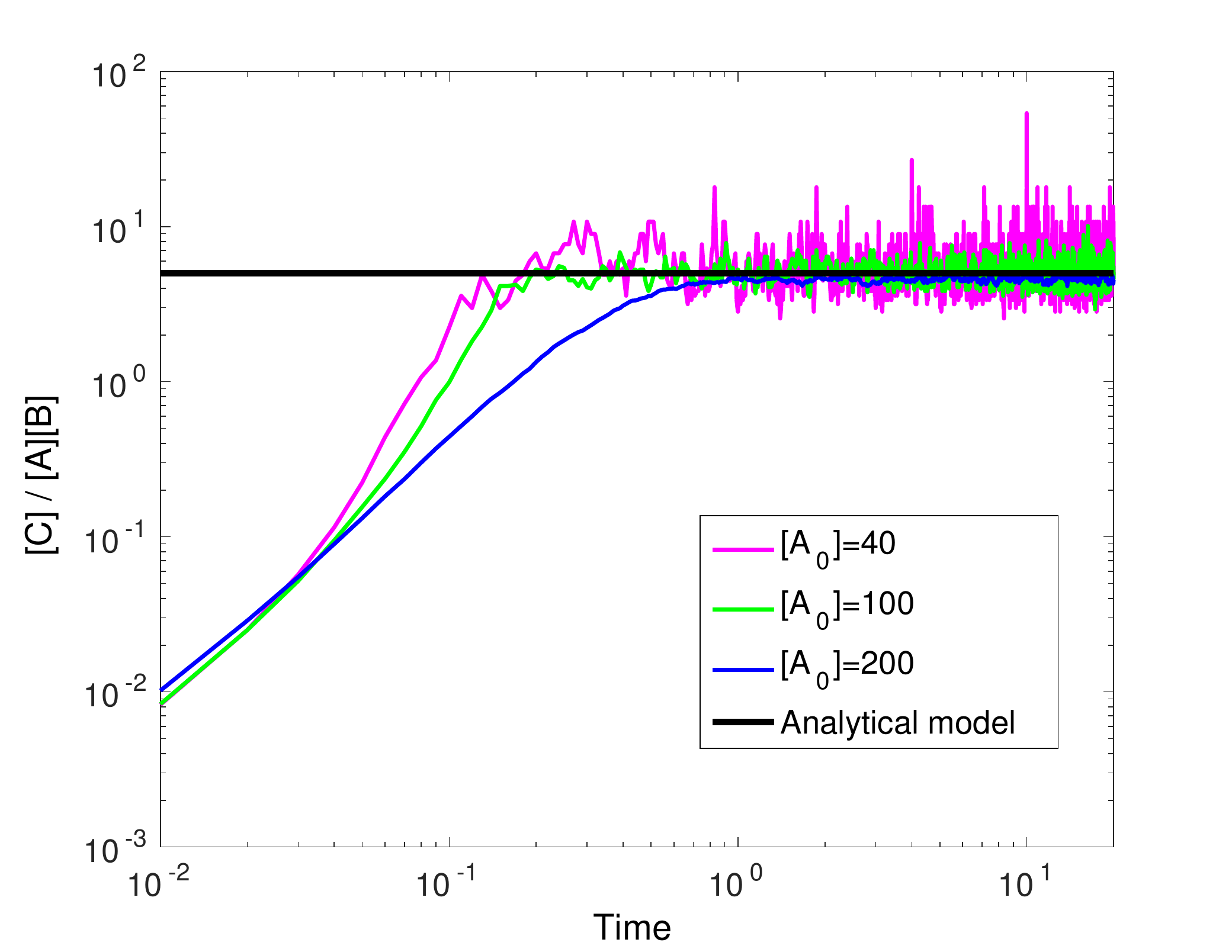}
    \caption{Concentration ratio $[C]/[A][B]$ as a function of time for different initial concentration values $[A_0]$. At equilibrium this ratio should be equal to $K_{eq}$, which is plotted for comparison purposes.}
  \label{fig:Keq}
\end{figure}

Figure~\ref{fig:Keq} shows the concentration ratio $[C]/[A][B]$ as a function of time for different initial concentration values $[A_0]$. It can be observed that all the simulations approach equilibrium at late times. Fluctuations around the equilibrium value decrease as $[A_0]$ increases. For a concentration of $[A_0] = 200$, which corresponds to an initial 40,000 particles, 
the actual equilibrium constant is closely approximated by the model at late times while the number of remaining adsorbate particles $A$ in the system is very low, in the order of 1000 particles. 
This reasonable approximation shows the potential of our proposed model to simulate adsorption with a low number of particles. We note that the classical diffusion-based model is unable to reach this approximation due to the segregation of particles at late times \citep{Rahbar2015}.

\begin{figure}[!htp]
  \includegraphics[width=30pc]{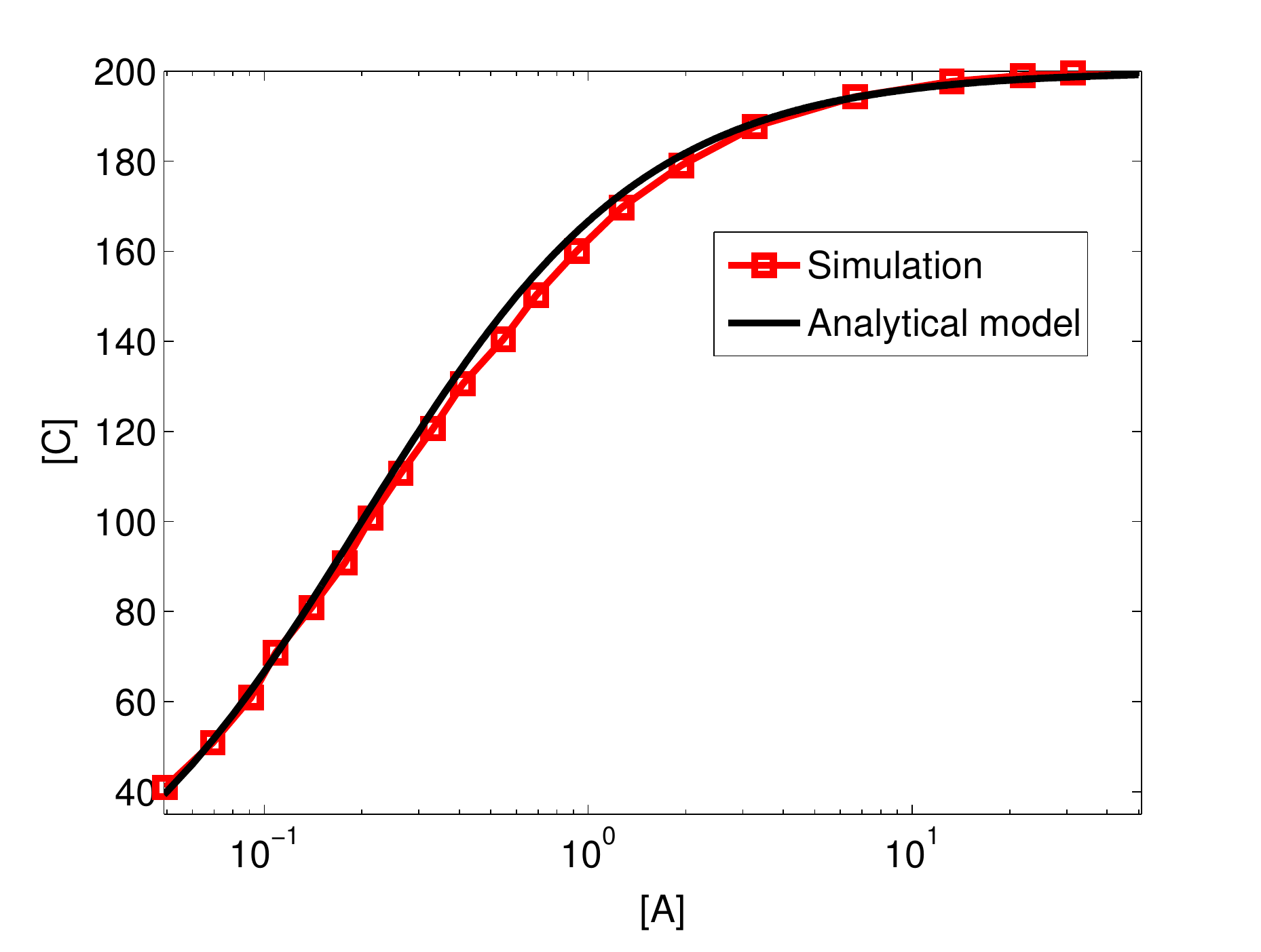}
  \caption{Concentration of the adsorbed species $[C]$ as a function of adsorbate concentration $[A]$ at equilibrium. Analytical solution of the Langmuir isotherm in Eq.~\eqref{Lang} is also plotted for comparison purposes.}  
  \label{fig:Lang}
\end{figure}

To further examine the performance of the proposed model, we performed twenty simulations considering different initial concentration values $[A_0]$ = 40, 50, ..., 250. For each simulation, we obtained 
the concentrations $[A]$ and $[C]$ from the average of their values at the last 100 time steps (so that equilibrium was achieved).
Figure~\ref{fig:Lang} shows these concentrations together with the results of the analytical model in Eq.~\eqref{Lang} for comparison purposes. 
It is clear that the simulation results have a good agreement with the analytical Langmuir isotherm even for a concentration of $[A] < 0.5$, where the initial number of particles is 8000 and the number of remaining adsorbate particles is below 100.
By increasing the concentration $[A]$, the product concentration $[C]$ tends to saturate towards $[B_0]$ which is the maximum capacity of free sites for adsorption. This saturation level implies the mono-layer adsorption in the Langmuir isotherm. 

\subsection{The Freundlich Model}
\label{F2}
This is a challenging model due to the need to specify different equilibrium constants to each individual sorption site. We first set up all the parameters, except $k_f$, 
identical to that of the previous section, to be consistent with the simulations for the Langmuir model. Then, according to the theoretical developments for deriving Eq.~\eqref{Cm4}, the constant of forward adsorption $k_f$ can be 
obtained in terms of the Freundlich constant $m$ for each pairs of $A$ and $B$ particles by $k_f = k_b K_{min} (1 - \zeta)^{-1/m}$, where $K_{min}$ in the minimum equilibrium constant given by Eq.~\eqref{u-ODE} considering a relative deviation $\epsilon$ = $10^{-1}$ and $\zeta$ is 
a random number generated from a standard uniform distribution. Thus, different $m$ values were used in the simulations. Having these parameters, we followed the same procedures explained in the previous section to implement the forward and backward reactions.

\begin{figure}[!htp]
  \includegraphics[width=30pc]{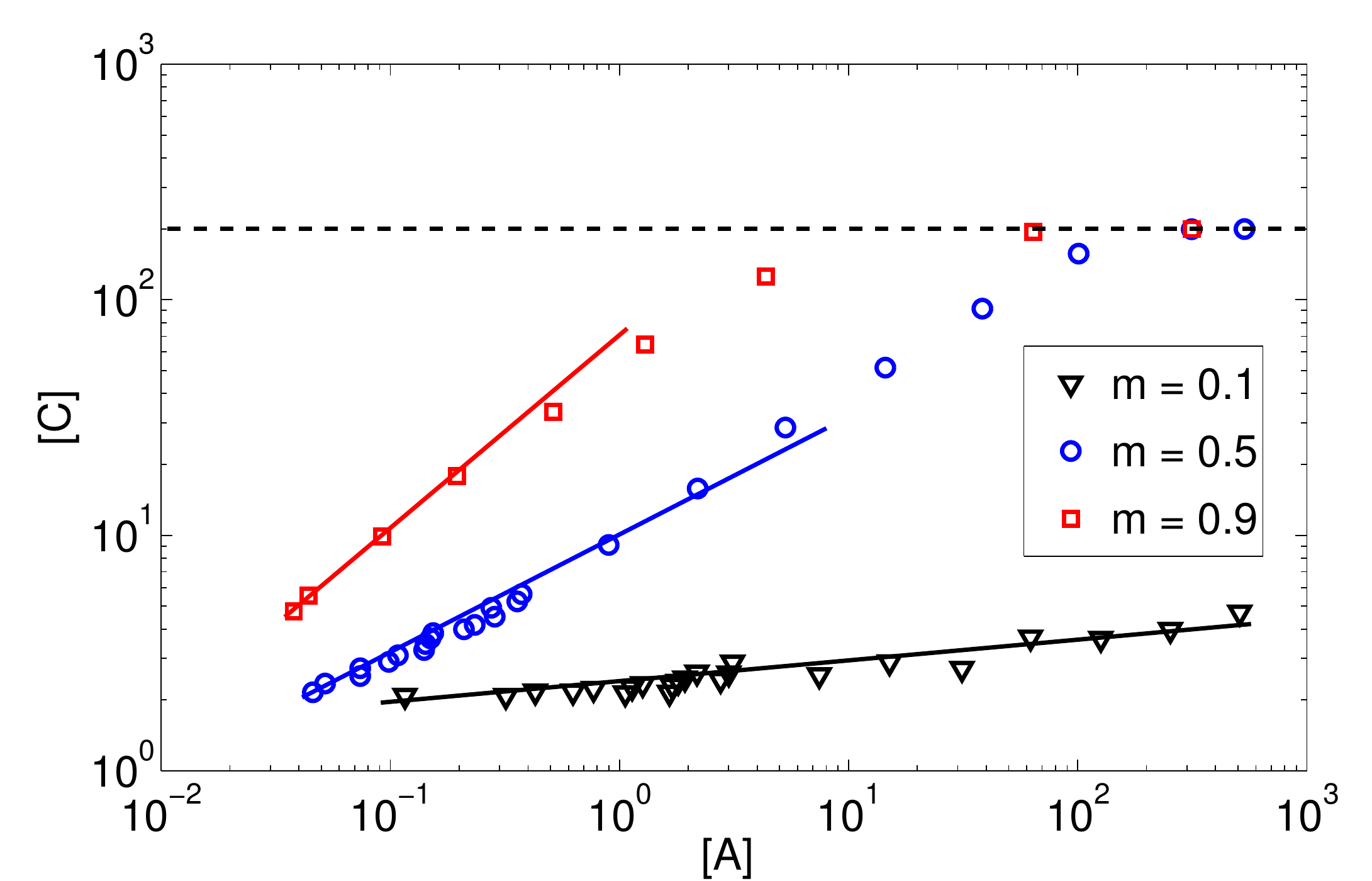}
  \caption{Concentration $[C]$ as a function of concentration $[A]$ at equilibrium. Logarithmic slope of the solid lines correspond to the Freundlich constant $m$ values used in the simulations. The dashed line shows the saturation concentration $[B_0]$.}  
  \label{fig:Fr}
\end{figure}

To obtain a good agreement between the model and the theory, the sorption sites should follow the exponential distribution of energy in Eq. \eqref{N}. However, in a range of low $[A]$ concentrations,
where the number of particles is small, there is a risk of undersampling. For this reason, we performed a larger number of simulations in this range. Figure~\ref{fig:Fr} shows the adsorbate concentration $[A]$ and the product concentration $[C]$ at equilibrium. It is clear that, in a range of low $[A]$ concentrations, 
the points in the graph follow closely a linear logarithmic slope whose value is controlled by the Freundlich constant $m$. By increasing the value of $m$, this agreement holds for a smaller range of concentrations $[A]$.

As mentioned in the introduction, a drawback of the Freundlich isotherm is that it cannot be employed for large adsorbate concentrations since this model does not taken into account the saturation limit of free adsorption sites. However, Fig.~\ref{fig:Fr} indicates that the proposed model is also able to incorporate this physical limitation without the need to use a complex model that transitions to a Langmuir-type behavior. Again in Figure~\ref{fig:Fr} it is observed that as the adsorbate concentration $[A]$ increases, the number of available adsorbent sites $B$ decreases and the product concentration $[C]$ tends to saturate to the maximum adsorbent concentration $[B_0]$, in agreement with Eq.~\eqref{integFrC8}. This saturation occurs at lower adsorbate concentration by increasing the value of $m$. Therefore, these results show that our model combines
the features of both the Langmuir and Freundlich isotherms in all range of adsorbate concentrations.

\section{Conclusions}\label{Concl}
Nonlinear adsorption in diffusion-reaction problems is a challenging problem. Eulerian methods lead to nonlinear partial differential equations, and so there is a need to develop efficient Lagrangian methods to tackle such a problem. While some methods have been proposed in the literature to address the Langmuir model, none so far is capable of addressing the Freundlich model. In this work we proposed a numerical method that combines a simple Particle Tracking Methods (PTM) with some predefined rules for particle interaction that properly reproduces nonlinear adsorption. The method uses Gaussian Kernel Density Estimators (KDEs), enabling it to avoid the effects of incomplete mixing due to the segregation of particles which is common in classical diffusion-based models with finite (and small) number of particles.  

For the Langmuir model, the method involves writing the adsorption process as the combination of a forward and a backward reaction. The former relies on the concept of particle distance, where one of the particles corresponds to the adsorbate (mobile) and the second one indicates a free site for adsorption (immobile). The probability of reaction is based on a function that involves the distance between both particles, and at each time step a random function is drawn to see if the reaction had taken place. The backward reaction is first-order. It was found that the PTM-KDE method provides a good reproduction of nonlinear adsorption following the Langmuir model by simply adding a finite number of sorption sites of homogeneous sorption properties. The advantage of using KDEs is that it is possible to obtain very good results in terms of adsorbed versus adsorbate concentrations with quite a small number of particles, while other PTM methods would rely on a very large number of tracked particles to obtain good approximations. The system quickly gets to equilibrium and it is very stable in terms of the very low number of particles that can be used. The effect of the nonlinearity resulted in a different equilibrium time depending on the initial concentration of the dissolved species (adsorbate). 

The approach was then extended to address nonlinear sorption modeled with a Freundlich isotherm. This involves some theoretical considerations regarding heterogeneity of the sorption properties at each specific sorption site. When such properties follow a truncated exponential statistical distribution, adsorption follows a power law in the ratio of sorbed and dissolved concentrations. This is valid for low to intermediate concentration values. 

The saturation of adsorbent sites has been also observed for high concentrations, pointing out the ability of our model to combine the features of the classical Langmuir and Freundlich isotherms. Our proposed approach opens up a new way to predict and control an adsorption-based process using a particle-based method with a finite (and actually quite low) number of particles.


\end{document}